\documentclass[sigconf, screen, nonacm]{acmart}
\AtBeginDocument{%
  }

\usepackage{xcolor}
\usepackage{makecell}
\usepackage{array}
\usepackage{colortbl}

\definecolor{mmhead}{RGB}{218,232,248}
\definecolor{mmsubhead}{RGB}{232,241,250}
\definecolor{mmmetric}{RGB}{242,248,253}
\definecolor{mmrowhi}{RGB}{250,244,230}

\begin{document}

\title{PolarMAE: Efficient Fetal Ultrasound Pre-training via Semantic Screening and Polar-Guided Masking}

\author{Meng Lv, Yapeng Li, Hang Su, Juhua Liu, Bo Du}
\affiliation{
  \institution{ Wuhan University}
  \city{Wuhan}
  \country{China}
}

\renewcommand{\shortauthors}{}
\begin{abstract}
Intelligent fetal ultrasound (US) interpretation is crucial for prenatal diagnosis, but high annotation costs and operator-induced variance make unsupervised pre-training a highly promising paradigm. However, existing pre-training methods largely ignore US-specific characteristics—severe data redundancy, fan-shaped locality, and polar coordinate beamforming—limiting their effectiveness in downstream tasks. To address this, we propose PolarMAE, a novel and efficient pre-training framework tailored for US images. Specifically, to mitigate continuous scanning redundancy, we introduce a Progressive Visual-Semantic Screening (PVSS) that adaptively extracts high-value samples, significantly boosting pre-training efficiency. Furthermore, we design an Acoustic-Bounded Region Constraint (ABRC) to accommodate US locality, forcing the model to focus strictly on valid acoustic regions rather than invalid dark backgrounds. Finally, leveraging the beamforming prior and local details, we propose a Polar-Texture Collaborative Masking (PTCM), enabling the model to capture underlying radial imaging patterns and critical tissue structures. Extensive experiments across diverse datasets and downstream interpretation tasks demonstrate that our method achieves state-of-the-art performance with strong pre-training scalability and efficiency.

\end{abstract}

\begin{teaserfigure}
  \centering
  \setlength{\fboxrule}{0.8pt}
  \setlength{\fboxsep}{0pt}
  \scalebox{1}[0.97]{%
    \fcolorbox{gray}{white}{%
      \includegraphics[width=0.992\textwidth]{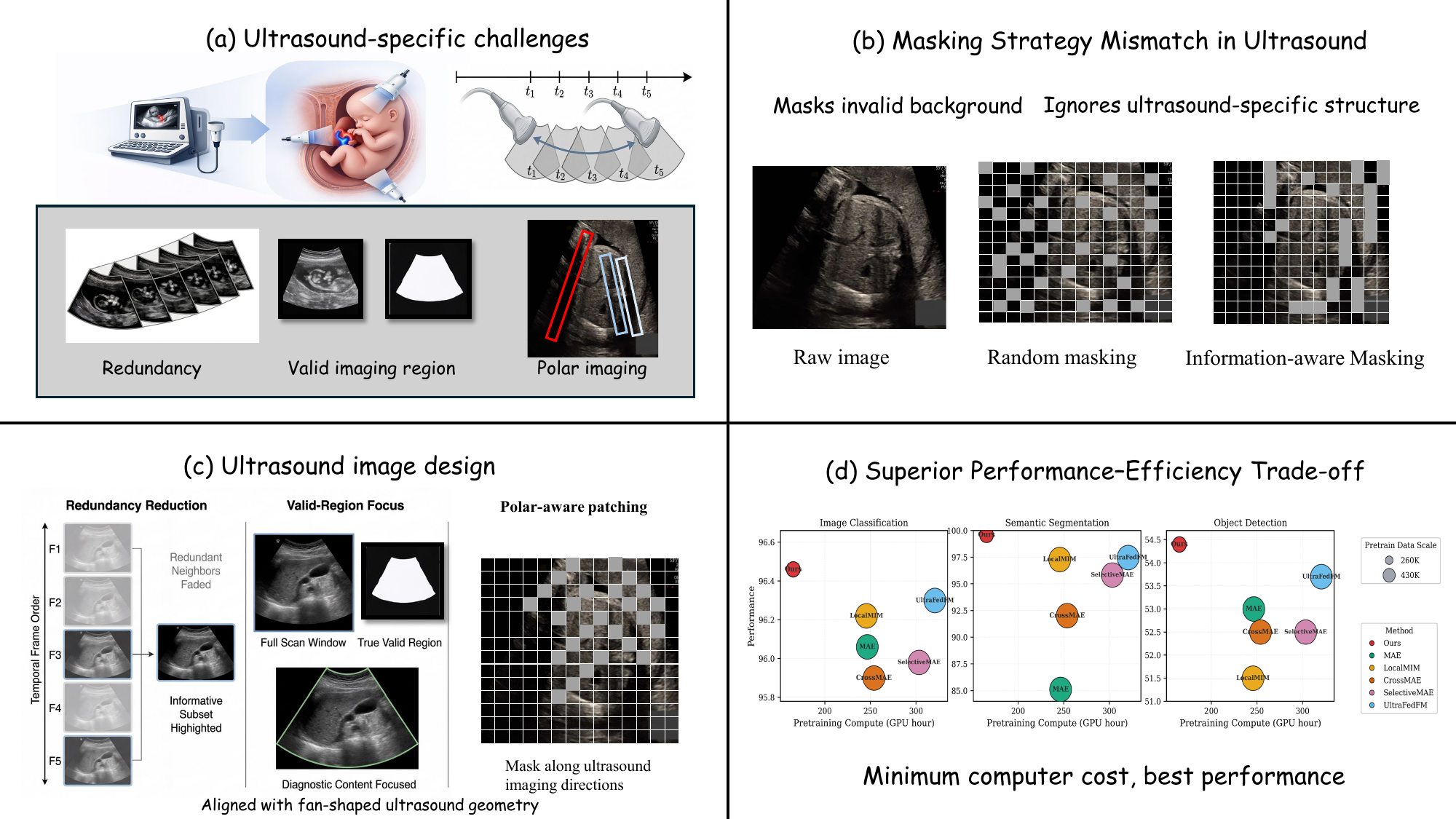}%
    }%
  }
  \vspace{-5pt}
  \caption{Motivation of PolarMAE. Generic MIM methods (b) mismatch the unique physical characteristics of ultrasound data (a), leading to inefficient pre-training. By customizing designs for redundancy reduction, valid-region focus, and polar-aware patching (c), our PolarMAE bridges this domain gap and achieves the best performance-efficiency trade-off with minimal computational cost across various interpretation tasks (d).}
  \vspace{-2pt}
  \label{fig:motivation}
\end{teaserfigure}
\maketitle

\section{Introduction}
Intelligent fetal ultrasound (US) interpretation has become increasingly crucial for modern prenatal diagnosis, offering a non-invasive, real-time, and cost-effective solution for various obstetric applications \cite{who1995ultrasound, who2013ultrasound, gao2023anatomically, cai2024lightweight}. In recent years, deep learning has driven significant progress in automated fetal US analysis \cite{fiorentino2023review, mhatre2024fetal, ramirez2023use}. However, developing robust models faces a critical bottleneck: the prohibitive cost of acquiring high-quality, pixel-level expert annotations, coupled with the severe variance in image distribution induced by different operators and scanning devices \cite{sendra2023generalisability, vega2025overcoming}. These fundamental challenges make unsupervised pre-training a highly promising paradigm, as it allows models to learn generic, transferable representations directly from massive amounts of unannotated clinical data \cite{shen2021miscell, kang2023deblurring, megahed2025usf}.

Recently, Masked Image Modeling (MIM) has emerged as a dominant unsupervised pre-training approach in visual representation learning \cite{he2022masked, xie2022simmim, bao2022beit}. Naturally, preliminary attempts have been made to extend MIM to US images to alleviate the annotation burden \cite{deblurringmae2023, xu2024maskedus, sslusseg2023, he2025masked}. However, existing pre-training methods, which are primarily designed for natural images, largely ignore US-specific data and physical characteristics \cite{kang2024deblurring, kang2025d}, fundamentally limiting their effectiveness and efficiency in downstream tasks(see Fig.~\ref{fig:motivation}b). Specifically, this domain gap manifests in three critical aspects(see Fig.~\ref{fig:motivation}a). First, \textbf{severe data redundancy}: clinical US examinations typically involve continuous video stream scanning  \cite{sharma2021knowledge}, resulting in massive temporal and semantic redundancy. Treating all frames equally wastes substantial computational resources on repetitive patterns rather than informative features. Second, \textbf{fan-shaped locality}: unlike the full-view rectangular framing of natural images, US imaging is typically confined to a specific fan-shaped local field of view. Generic random masking strategies inevitably force the model to reconstruct invalid, dark background regions. Third, \textbf{polar coordinate beamforming}: dictated by the physics of acoustic beamforming, US spatial information follows a non-uniform, radial distribution. Standard pre-training methods treat the image as a uniform Cartesian grid, completely overlooking this underlying radial imaging pattern and failing to prioritize critical tissue structures.

To address these critical limitations, we propose PolarMAE, a novel and highly efficient pre-training framework explicitly tailored for ultrasound images (see Fig.~\ref{fig:motivation}c). PolarMAE rethinks the masked autoencoding paradigm by deeply integrating acoustic physics and US data properties. Specifically, to mitigate the severe visual redundancy caused by continuous scanning, we introduce a Progressive Visual-Semantic Screening (PVSS). Unlike generic MAE that processes all frames indiscriminately, this mechanism adaptively evaluates and extracts high-value, non-redundant samples from massive US sequences across both structural and semantic levels. This selective process significantly boosts pre-training efficiency and prevents the model from overfitting to repetitive, low-information patterns. Furthermore, to accommodate the unique fan-shaped locality of US data, we design an Acoustic-Bounded Region Constraint (ABRC). Instead of applying random masking across the entire Cartesian grid, this strategy explicitly defines the acoustic boundaries, forcing the model to focus its reconstruction efforts strictly on valid anatomical regions and eliminating wasteful gradient updates on invalid dark backgrounds. Finally, leveraging the inherent physical prior of acoustic beamforming, we propose a Polar-Texture Collaborative Masking (PTCM) that jointly models macroscopic geometry and microscopic details. By aligning the masking and representation learning process with the radial distribution of US signals and local textures, this strategy enables the model to actively capture the underlying radial imaging patterns and prioritize the learning of critical tissue structures.

The main contributions of this work are summarized as follows:
\begin{itemize}
    \item We propose PolarMAE, a highly efficient unsupervised pre-training framework customized for fetal ultrasound. It fundamentally bridges the gap between generic MIM and US-specific acoustic characteristics.
    \item We introduce a Progressive Visual-Semantic Screening (PVSS) to filter redundant frames, coupled with an Acoustic-Bounded Region Constraint (ABRC). These designs drastically accelerate pre-training and eliminate invalid background computation.
    \item We design a Polar-Texture Collaborative Masking (PTCM) that injects the physical prior of acoustic beamforming and local texture into the representation learning process, effectively capturing non-uniform radial imaging patterns and structural details.
\end{itemize}

Extensive experiments across diverse datasets and downstream interpretation tasks demonstrate that PolarMAE achieves state-of-the-art performance (see Fig.~\ref{fig:motivation}d). It significantly accelerates pre-training while exhibiting exceptional representation quality and scalability.

\section{Related Work}
\label{sec:related_work}
\subsection{Masked Image Modeling}
\label{subsec:rw_ssl_mim}

In recent years, self-supervised learning has become an essential technique for visual representation learning, especially in scenarios where large-scale unlabeled data are available but manual annotation is expensive and difficult to obtain \cite{chen2020simple, he2020momentum}. Among different self-supervised paradigms, Masked Image Modeling (MIM) has attracted particular attention due to its simple formulation and strong scalability \cite{jing2020self}. The core idea of MIM is to mask part of the input and train the model to infer the missing content from the visible context, thereby learning robust and transferable visual representations without human supervision. Existing MIM methods are generally grouped into reconstruction-based and contrastive-based frameworks, among which reconstruction-based methods have become the dominant line in practice \cite{zhou2021ibot, wei2022masked, baevski2022data2vec, tao2023siamese}. Representative studies such as BEiT \cite{bao2022beit}, MAE \cite{he2022masked} and SimMIM \cite{xie2022simmim} establish a simple yet scalable pre-training paradigm by masking a large portion of image patches and recovering the missing content from the visible context. Building on this formulation, subsequent studies further extend MIM from raw-pixel reconstruction to latent-token prediction and semantic-token reconstruction\cite{chang2022maskgit, li2023mage}, and further broaden it to multimodal masked modeling\cite{fan2024text, guo2024crossmae}, demonstrating strong transferability across classification, segmentation, and generation tasks.

These developments indicate that  MIM gradually develops from a basic masked reconstruction framework into an important paradigm for visual self-supervised pretraining, and this progress provides a foundation for its further application to medical imaging scenarios with complex structural priors.

\subsection{Masked Image Modeling in Medical Imaging}
\label{subsec:rw_medical_ssl}
Medical imaging is one of the most important application domains for self-supervised learning, since clinical data are often abundant while expert annotations are expensive, time-consuming, and sometimes difficult to standardize across institutions. Therefore, self-supervised pre-training has attracted increasing attention for its ability to learn transferable representations from large-scale unlabeled data, providing a highly effective solution to the annotation bottleneck in medical imaging. \cite{zhou2023review, wang2024devil, phutke2024oscmamba, zhou2024s}. For example, MAESTER \cite{xie2023maester} incorporates masked autoencoding into a vision transformer pipeline to learn token representations for sub-cellular structure segmentation, while AMAE \cite{bozorgtabar2023amae} leverages MAE-style masked reconstruction for chest X-ray anomaly detection. Cai et al. further design a unified patch embedding framework for both 2D and 3D ophthalmic images \cite{cai2022uni4eye}, and Chen et al. employ MIM to learn latent representations from fMRI data for downstream diffusion-based visualization \cite{chen2023masked}. More recently, MIM has been further extended to volumetric medical imaging through local-global masked reconstruction combined with teacher-student optimization \cite{zhang2024mapseg}. Collectively, these studies demonstrate the value of MIM for exploiting large-scale unlabeled clinical data and improving downstream performance under limited supervision.

However, most existing medical MIM methods still largely follow a generic transfer paradigm, with only limited adaptation to modality-specific characteristics. As a result, factors such as acquisition geometry, valid imaging regions, and structured redundancy are often insufficiently modeled during pre-training. This suggests that further improvements may depend on better integrating modality-specific priors into the masking and reconstruction process.

\subsection{Masked Image Modeling in Ultrasound Imaging}
\label{subsec:rw_ultrasound_ssl}

Ultrasound imaging is widely used in clinical practice because it is radiation-free, real-time, portable, and relatively low-cost. Its broad clinical adoption also results in large amounts of unlabeled data, making ultrasound a favorable setting for self-supervised pre-training. Existing studies have shown that self-supervised learning can improve ultrasound image analysis under limited annotation and benefit downstream tasks such as classification and segmentation \cite{sslusseg2023, wang2024devil, rahman2024ultramae, kang2025d}. For example, Xu et al. apply random masking to ultrasound image classification and show that generic masked modeling can already learn useful ultrasound representations \cite{xu2024maskedus}. Subsequent studies further adapt pre-training objectives to ultrasound characteristics. Kang et al. introduce a deblurring objective to improve fine-grained structural representation under blur and noise \cite{deblurringmae2023}. USFM, proposed by Jiao et al., introduces a spatial-frequency dual masked image modeling strategy on a large-scale multi-organ, multi-center, and multi-device ultrasound dataset \cite{jiao2024usfm}. UltraFedFM, proposed by Jiang et al., further combines self-supervised pre-training with federated learning to train an ultrasound foundation model on distributed multi-institutional data \cite{jiang2025ultrafedfm}.

Overall, existing studies confirm the potential of MIM for ultrasound image analysis. However, most methods still adopt relatively generic masking and reconstruction strategies and do not fully model ultrasound-specific properties such as sequence redundancy, valid imaging regions, and imaging geometry. This leaves substantial room for a pre-training framework that more deeply incorporates ultrasound-specific data structure and physical priors.

\begin{figure*}[t]
  \centering
  \includegraphics[width=\textwidth]{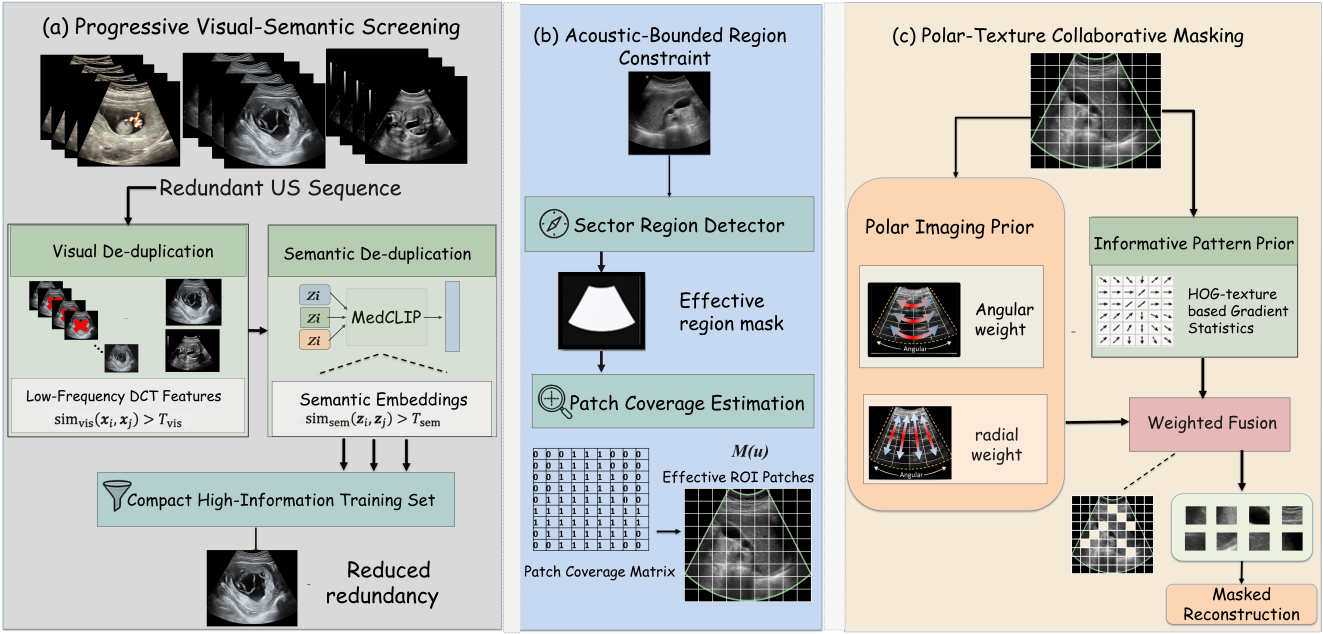}
  \caption{Overview of the proposed PolarMAE framework. Our pre-training method consists of three stages. First, semantic de-duplication uses visual and MedCLIP filters to eliminate redundant frames. Second, effective region extraction explicitly discards invalid background patches. Finally, the collaborative masking strategy fuses polar geometric priors and HOG texture responses to guide targeted mask sampling. This US-tailored pipeline enables the model to efficiently learn robust representations from structurally informative tissue regions.}
  \vspace{-13pt}
  \label{fig:framework}
\end{figure*}

\section{Method}
\label{sec:method}
\subsection{Motivation and Overall Idea}
\label{subsec:method_overview}

Existing self-supervised pretraining methods for ultrasound images usually follow generic masked modeling paradigms developed for natural images and do not fully exploit the polar-coordinate physical characteristics of beam imaging. As a result, both representation quality and downstream generalization still have room for improvement. Moreover, when applied to large-scale ultrasound data, standard pretraining pipelines often model many redundant samples and invalid background regions, which leads to substantial computational and time overhead. To address these issues, we propose a self-supervised pretraining method for ultrasound images that improves both representation quality and pretraining efficiency. Built on the MAE framework, the proposed method refines the pretraining pipeline from three aspects: data selection, region constraint, and spatial prior modeling.

Specifically, PolarMAE is driven by a physics-informed design philosophy. We first introduce Progressive Visual-Semantic Screening (PVSS) to filter continuous scanning redundancy from both visual and semantic perspectives. Then, to accommodate the unique fan-shaped locality, an Acoustic-Bounded Region Constraint (ABRC) is applied to focus computational resources exclusively on valid areas. Finally, within this bounded region, we propose Polar-Texture Collaborative Masking (PTCM), which seamlessly integrates the macroscopic geometry of acoustic beamforming with microscopic tissue texture to guide patch selection.

\subsection{Masked Autoencoders Preliminaries}
\label{subsec:mae_baseline}

Our method is built upon the Masked Autoencoder (MAE) framework \cite{he2022masked}. In MAE, an input image is first divided into fixed-size non-overlapping patches. According to a predefined masking ratio, part of the patches is randomly retained as visible tokens, while the remaining patches are masked. The encoder models context only from the visible tokens, and the decoder reconstructs the masked regions by combining the encoded visible features with mask tokens. Standard MAE uses pixel-level reconstruction as the pretraining objective and measures the difference between the predicted and original patches with mean squared error:
\begin{equation}
\mathcal{L}_{rec}=\frac{1}{|M|}\sum_{i\in M}\|\hat{x}_i-x_i\|_2^2,
\end{equation}
where $M$ denotes the index set of masked patches.

Standard MAE learns contextual dependencies through random masking and masked-region reconstruction, and has shown strong effectiveness in visual self-supervised pretraining \cite{he2022masked}. However, it implicitly assumes that different patches have roughly equal importance in encoding and reconstruction, which is not fully suitable for ultrasound images. Because ultrasound images often contain redundancy from continuous scanning, large invalid black background regions, and polar-coordinate spatial characteristics determined by beam imaging, the random masking and uniform reconstruction strategy of standard MAE cannot fully balance representation quality and pretraining efficiency in this setting. Therefore, based on the MAE baseline, we introduce targeted improvements from three aspects: data de-redundancy, effective-region-constrained reconstruction, and polar-prior-guided modeling.

\subsection{Progressive Visual-Semantic Screening}
\label{subsec:dual_stage_dedup}
Standard MAE indiscriminately processes all input frames, ignoring the severe temporal and semantic redundancy inherent to continuous clinical ultrasound scanning. Adjacent frames in US video streams often exhibit high visual similarity. Furthermore, even images with varying appearances due to slight probe movements can share highly consistent medical semantics. Directly pre-training on such uncurated data forces the model to expend substantial computational resources learning repetitive distributions, severely diluting the information density of the training corpus. To overcome this, we propose the Progressive Visual-Semantic Screening (PVSS) module, a hierarchical data curation strategy that filters redundant samples across both structural and semantic dimensions.

In the first phase of PVSS, we perform visual-level screening to eliminate temporally adjacent near-duplicates. We extract low-frequency Discrete Cosine Transform (DCT) features to robustly represent the global structural information of each frame, effectively bypassing high-frequency noise. The visual similarity between two samples is measured via cosine similarity:
\begin{equation}
\mathrm{sim}_{vis}(x_i,x_j)=
\frac{f_i^\top f_j}{\|f_i\|\,\|f_j\|},
\end{equation}
where $f_i$ and $f_j$ denote the low-frequency structural features of images $x_i$ and $x_j$, respectively. For samples whose visual similarity exceeds a predefined threshold, only one sample is retained.

While filtering effectively removes visually identical frames, it struggles to identify semantic redundancy caused by viewpoint shifts. Therefore, in the second phase, we map the remaining images into a high-level semantic space. Specifically, we leverage the pre-trained image encoder of MedCLIP to extract semantic embeddings and compute their cosine similarity:

\begin{equation}
\mathrm{sim}_{sem}(x_i,x_j)=
\frac{z_i^\top z_j}{\|z_i\|\,\|z_j\|},
\end{equation}

where $z_i$ and $z_j$ represent the semantic embeddings of the corresponding images. Samples with $\mathrm{sim}_{sem}$ exceeding a predefined semantic threshold are treated as semantically redundant and discarded. It is crucial to emphasize that MedCLIP is utilized strictly as an offline dataset curation tool. Its weights remain completely frozen, and no supervised signals or gradients are back-propagated into the subsequent PolarMAE pre-training process, thereby preserving the self-supervised nature of our framework.

Through this progressive visual and semantic filtering, PVSS effectively distills massive, redundant ultrasound sequences into a highly compact training set. This hierarchical screening drastically increases the overall information density, enabling PolarMAE to learn rich representations with significantly reduced pre-training overhead.

\subsection{Acoustic-Bounded Region Constraint}
\label{subsec:sector_region_extraction}

Beyond temporal and semantic redundancy, standard MAE paradigms treat the entire Cartesian image grid uniformly, failing to distinguish the effective acoustic imaging region from the invalid background. Dictated by the physics of ultrasound probes, US images typically exhibit a bounded, fan-shaped field of view (FOV), leaving substantial dark background regions devoid of anatomical information. If these invalid regions are masked and reconstructed uniformly alongside true tissue areas, the model inevitably squanders precious computational budget and representation capacity on zero-information pixels. To overcome this, we introduce the Acoustic-Bounded Region Constraint (ABRC) to explicitly confine the learning objective to valid anatomical areas.

Specifically, the corresponding sector-region mask is defined as
\begin{equation}
M(u)=
\begin{cases}
1, & u\in\Omega_{\text{sector}},\\
0, & \text{otherwise},
\end{cases}
\end{equation}
where $\Omega_{\text{sector}}$ denotes the detected effective imaging region in the ultrasound image, and $M(u)$ is the corresponding binary mask. Based on this mask, an ROI mask is generated for subsequent patch-level region modeling.

Furthermore, in the subsequent patch-level modeling stage, given the pixel set $\mathcal{P}_i$ corresponding to the $i$-th patch, its coverage ratio within the effective sector region is defined as
\begin{equation}
v_i=\frac{1}{|\mathcal{P}_i|}\sum_{u\in\mathcal{P}_i} M(u).
\end{equation}
This quantity characterizes the extent to which a patch falls inside the effective sector region and serves as the basis for the subsequent polar-coordinate geometric prior and joint probabilistic sampling.

Through this process, the effective imaging region of each ultrasound image is explicitly identified before pre-training, which provides a more stable input basis for subsequent Polar-Texture Collaborative Masking guided by the ultrasound imaging prior.

\begin{table*}[!t]
\centering
\caption{Comparison of different pre-training methods on fetal ultrasound downstream tasks. Best results are shown in bold. PUBSEG denotes the public benchmark used for semantic segmentation evaluation.}
\label{tab:pretrain_comparison}
\small
\renewcommand{\arraystretch}{1.10}
\setlength{\tabcolsep}{0pt}
\setlength{\aboverulesep}{0pt}
\setlength{\belowrulesep}{0pt}
\setlength{\cmidrulesep}{0.1ex}

\begin{tabular}{@{}
    >{\centering\arraybackslash}p{0.13\textwidth}
    >{\centering\arraybackslash}p{0.12\textwidth}
    >{\centering\arraybackslash}p{0.10\textwidth}
    >{\centering\arraybackslash}p{0.11\textwidth}
    >{\centering\arraybackslash}p{0.10\textwidth}
    >{\centering\arraybackslash}p{0.11\textwidth}
    >{\centering\arraybackslash}p{0.11\textwidth}
    >{\centering\arraybackslash}p{0.11\textwidth}
    >{\centering\arraybackslash}p{0.11\textwidth}
@{}}
\toprule
\rowcolor{mmhead}
\multicolumn{3}{c}{} &
\multicolumn{2}{c}{\textbf{Image Classification}} &
\multicolumn{2}{c}{\textbf{Semantic Segmentation}} &
\multicolumn{2}{c}{\textbf{Object Detection}} \\
\cmidrule{4-9}

\rowcolor{mmsubhead}
\textbf{Model} &
\textbf{Venue} &
\textbf{Pretrain Data} &
\textbf{Private} &
\textbf{SFP} &
\textbf{Private} &
\textbf{PUBSEG} &
\textbf{Private} &
\textbf{FIS} \\
\cmidrule{4-5}
\cmidrule{6-7}
\cmidrule{8-9}

\rowcolor{mmmetric}
& & &
\textit{Acc} &
\textit{Acc} &
\textit{mDice} &
\textit{mDice} &
\textit{mAP@50:95} &
\textit{mAP@50:95} \\
\midrule
MAE           & CVPR'22            & 430K & 93.57          & 96.06          & 71.27          & 85.11          & 63.0          & 53.0 \\
LocalMIM      & CVPR'23            & 430K & 92.95          & 96.22          & 74.82          & 97.29          & 63.7          & 51.5 \\
SelectiveMAE  & ICCV'25            & 430K & 93.69          & 95.98          & 73.57          & 95.83          & 62.5          & 52.5 \\
CrossMAE      & TMLR'25            & 430K & 93.85          & 95.90          & 75.67          & 92.02          & 64.1          & 52.5 \\
UltraFedFM    & npj Digit. Med.'25 & 430K & 94.02          & 96.30          & 72.35          & 97.47          & 63.4          & 53.7 \\
\rowcolor{mmrowhi}
PolarMAE        & --                 & 260K & \textbf{94.34} & \textbf{96.46} & \textbf{77.82} & \textbf{99.59} & \textbf{65.6} & \textbf{54.4} \\
\bottomrule
\end{tabular}
\end{table*}

\subsection{Polar-Texture Collaborative Masking}

Following the application of PVSS and ABRC, the representation learning is effectively constrained to valid anatomical regions. However, dictated by the acoustic beamforming mechanism, the spatial distribution of information within this fan-shaped FOV remains highly non-uniform. Standard uniform random masking fails to exploit this underlying physical characteristic. To address this, we propose Polar-Texture Collaborative Masking (PTCM), which constructs a joint probability distribution to guide mask patch selection and reconstruction target sampling by synergizing beamforming-induced polar priors with texture details.

Specifically, patches with $v_i\ge\tau$ are regarded as belonging to the effective ROI, where $\tau$ is the threshold for effective coverage.
Based on this, a polar-coordinate geometric prior is constructed from the connected ROI region on the patch grid.
Specifically, for a patch centered at grid coordinate $(m_i,n_i)$, we estimate the top row, bottom row, and centerline of the connected ROI component as $m_{\mathrm{apex}}$, $m_{\mathrm{bottom}}$, and $n_{\mathrm{center}}$, respectively. Let $h(m_i)$ denote the half-width of the connected ROI at row $m_i$. The normalized radial position and angular offset are then defined as
\begin{equation}
r_i=\mathrm{clip}\!\left(\frac{m_i-m_{\mathrm{apex}}}{m_{\mathrm{bottom}}-m_{\mathrm{apex}}},\,0,\,1\right),
\end{equation}
\begin{equation}
\theta_i=\mathrm{clip}\!\left(\frac{n_i-n_{\mathrm{center}}}{h(m_i)},\,-1,\,1\right),
\end{equation}
where $r_i$ measures the normalized depth of the patch within the connected sector region, and $\theta_i$ measures its normalized lateral deviation from the sector centerline.
The radial weight, angular weight, and coverage gating term are defined as
\begin{equation}
f_r(i)=\exp\!\left(-\frac{(r_i-\mu)^2}{2\sigma^2}\right),
\end{equation}
\begin{equation}
g_{\theta}(i)=1-|\theta_i|^k,
\end{equation}
\begin{equation}
q_i=\mathrm{clip}\!\left(\frac{v_i-\tau}{1-\tau},\,0,\,1\right),
\end{equation}
where $\mu$ and $\sigma$ control the location and width of the radially high-response region, $k$ is the angular decay coefficient, and $\mathrm{clip}(x,0,1)=\min(\max(x,0),1)$ truncates the input to $[0,1]$. The gating term $q_i$ maps each patch's ROI coverage to a continuous weight, softly modulating the polar-coordinate geometric score to suppress patches near the ROI boundary while emphasizing patches well within the effective region. The polar-coordinate geometric score is then defined as
\begin{equation}
s_i^{polar}=q_i\left(f_r(i)+g_{\theta}(i)\right),
\end{equation}
and the corresponding prior distribution is obtained by normalization:
\begin{equation}
p_i^{polar}=\frac{s_i^{polar}}{\sum_j s_j^{polar}}.
\end{equation}

To further characterize local texture strength, we introduce the histogram of oriented gradients (HOG) response. Let $s_i^{hog}$ denote the HOG score of the $i$-th patch. Its texture probability distribution is defined as
\begin{equation}
p_i^{hog}=\frac{\exp(s_i^{hog})}{\sum_j \exp(s_j^{hog})}.
\end{equation}
The two distributions are then fused to obtain the joint probability
\begin{equation}
p_i=(1-\lambda)p_i^{hog}+\lambda p_i^{polar},
\end{equation}
where $\lambda\in[0,1]$ is the balancing coefficient between texture response and geometric prior.

During training, $p_i$ is used for both visible token selection and reconstruction target sampling. A candidate set is first built from effective ROI patches. When the number of effective patches is insufficient, additional patches are supplemented from outside the ROI to maintain a stable token count within each batch. Initial visible tokens are then sampled according to the joint probability, and the encoder input is progressively expanded by a semantic token selection strategy. Finally, the reconstruction target set $\mathcal{T}_{recon}$ is sampled from the remaining unselected effective patches using the same joint probability, where $\mathcal{T}_{recon}$ denotes the set of target patches reconstructed by the decoder.

Through this design, the patch selection process jointly considers the geometric structure of ultrasound imaging and local texture information, enabling the model to focus on more informative regions within the effective field of view and thereby improving both pre-training efficiency and representation quality.

\section{Experiments}
\label{sec:experiments}
In this section, we conduct a series of comprehensive experiments to evaluate the effectiveness and efficiency of the proposed PolarMAE framework. Our evaluation is centered around the following four research questions:
\begin{itemize}
\item \textbf{RQ1 (Cross-Task Versatility)}: Can PolarMAE learn generic representations for diverse ultrasound requirements? Does its versatility hold across a task spectrum of increasing granularity, from image-level classification to pixel-level segmentation?

\item \textbf{RQ2 (Design Synergy)}: How do PVSS, ABRC, and PTCM individually contribute to the overall gains? Does the specific polar-guided mask design effectively facilitate the capture of ultrasound-inherent imaging patterns?

\item \textbf{RQ3 (Computational Efficiency)}: Does PolarMAE strike a superior balance between computational economy and representation quality? Can it accelerate pre-training while yielding competitive performance?

\item \textbf{RQ4 (Data Scalability)}: Does the framework exhibit favorable scaling properties as data volume increases? Can it consistently translate more unannotated data into downstream gains?
\end{itemize}

\begin{figure}[!t]
  \centering
  \includegraphics[width=\columnwidth]{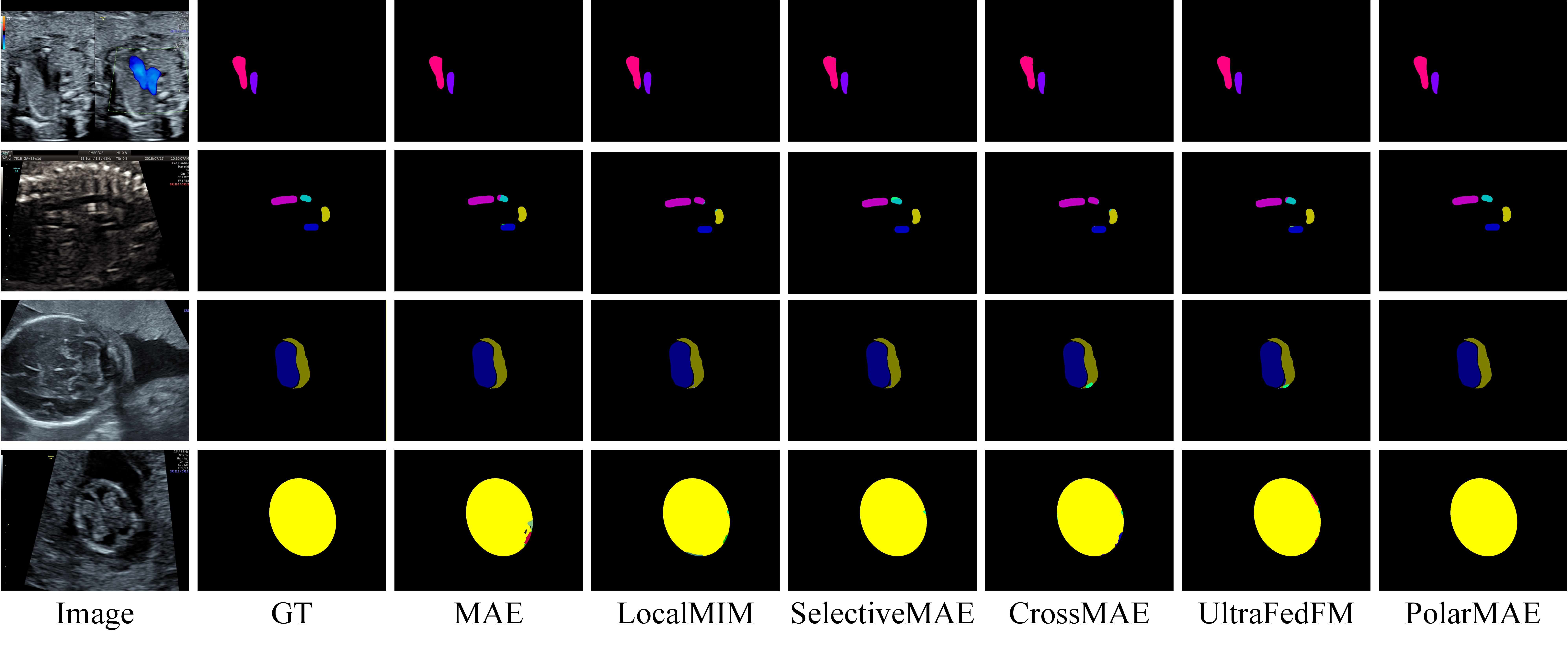}
  \caption{Qualitative results on private downstream segmentation.}
  \Description{Qualitative comparison of downstream segmentation results.}
  \label{fig:seg_vis}
\end{figure}

\subsection{Experimental Setting}
\noindent\textbf{Datasets.} We evaluate PolarMAE using a combination of large-scale clinical data and diverse downstream benchmarks.
\begin{itemize}
\item \textbf{Pre-training Dataset:} To ensure the robustness of representation learning, we construct a large-scale, multi-center fetal ultrasound dataset. It comprises approximately 430,000 frames (reduced to 260,000 high-information frames via our PVSS module) collected from multiple clinical centers with diverse imaging devices, ensuring a broad coverage of acoustic appearances.
\item \textbf{Downstream Benchmarks:} To validate the versatility of PolarMAE, we cover a multi-granularity task spectrum. Specifically, we utilize public benchmarks including SFP (classification) \cite{zenodo_classification}, FIS (detection) \cite{mendeley_detection}, and PUBSEG (segmentation). Recognizing the limited diversity in existing public datasets, we further construct three corresponding private clinical datasets to provide a more comprehensive assessment of generalization under real-world scenarios.
\end{itemize}
Detailed data construction and comparison metrics are provided in the Supplementary Material.


\noindent\textbf{Implementation Details.} PolarMAE is implemented in PyTorch and executed on four NVIDIA RTX 4090 GPUs. To support reproducibility, our complete codebase and logs will be made open-source upon acceptance. Specific implementation details for pre-training and various downstream tasks are provided in the Supplementary Material.

\begin{figure}[!t]
  \centering
  \includegraphics[width=\columnwidth]{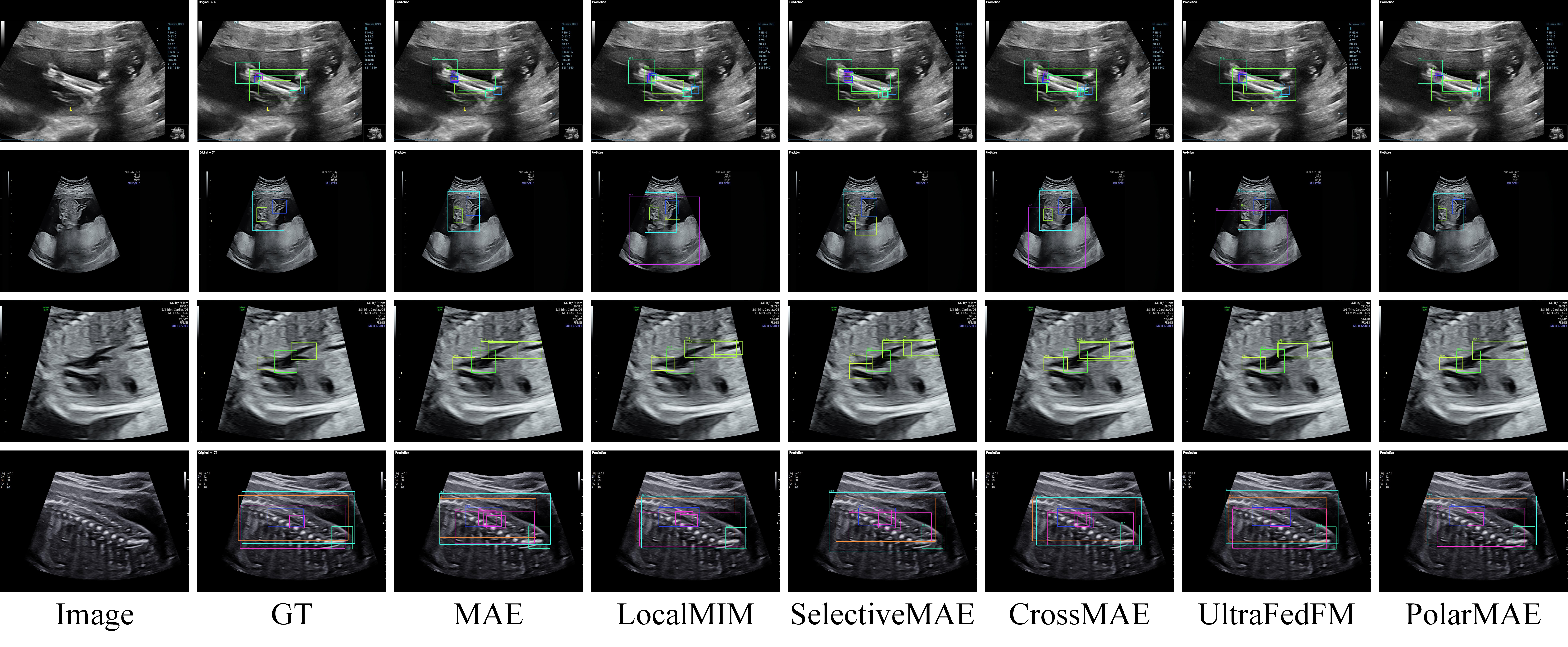}
  \caption{Qualitative results on downstream object detection.}
  \Description{A single-column placeholder figure for downstream object detection visualization.}
  \label{fig:det_vis}
\end{figure}
\subsection{Comparison with State-of-the-Arts (RQ1)}
\label{subsec:main_results}

We compare PolarMAE with several representative pre-training methods, including classical MIM methods such as MAE~\cite{he2022masked} and LocalMIM~\cite{Wang_2023_CVPR}, recent self-supervised methods such as SelectiveMAE~\cite{Wang_2025_ICCV} and CrossMAE~\cite{fu2025rethinking}, and the ultrasound-specific pre-training method UltraFedFM~\cite{jiang2025ultrafedfm}. Unlike the competing methods, which are pretrained on the original dataset, PolarMAE uses a compact subset produced by dual-stage filtering. Table~\ref{tab:pretrain_comparison} reports the results on image classification, object detection, and semantic segmentation. Despite the smaller pre-training set, PolarMAE achieves the best performance on all evaluated benchmarks, indicating that the proposed filtering strategy improves representation quality and downstream transfer while also reducing pre-training cost.

\textbf{Image Classification.}
On both the private classification dataset and the public SFP benchmark, PolarMAE achieves the best classification performance. This result indicates that the representations learned by PolarMAE transfer well across data sources rather than adapting only to a specific data distribution. It further suggests that suppressing redundancy in the pre-training data and directing the model toward regions with higher diagnostic value lead to more robust and more discriminative ultrasound representations.

\textbf{Object Detection.}
On the FIS detection task, PolarMAE again outperforms all competing methods. This result indicates that the representations learned by PolarMAE support not only image-level recognition but also reliable localization of key anatomical structures. It further suggests that the proposed method reduces interference from invalid background regions and strengthens the model's focus on diagnostically relevant areas.

\textbf{Semantic Segmentation.}
On the segmentation task, PolarMAE also achieves the best results on all evaluated benchmarks. This result indicates that PolarMAE is better suited to pixel-level structural modeling. Since semantic segmentation depends more heavily on local spatial relations and boundary information, the results suggest that the proposed method preserves structural contours and boundary cues in ultrasound images more effectively, thereby improving the representation of anatomical details.


\textbf{Qualitative Evaluation.}
From Fig~\ref{fig:seg_vis} and Fig~\ref{fig:det_vis}, we find that benefiting from the polar-guided masking strategy, PolarMAE produces significantly sharper and more anatomically consistent boundaries in complex fetal structures. Unlike generic Cartesian-based MIM, our model captures the underlying radial imaging patterns, ensuring structural continuity even in low-contrast regions.

\subsection{Ablation Study (RQ2)}
\subsubsection{Analysis of Key Module Contributions}
\label{subsubsec:module_ablation}

To evaluate the role of each key design, we conduct ablation studies on Dual-Stage Progressive Visual-Semantic Screening (PVSS), Acoustic-Bounded Region Constraint (ABRC), and Polar-Texture Collaborative Masking (PTCM), as shown in Table~\ref{tab:ablation_pretrain}. This experiment examines whether the three modules respectively alleviate sample redundancy, interference from invalid background regions, and non-uniform information distribution in ultrasound image pre-training.
\begin{table}[!t]
\centering
\caption{Ablation study on different masking strategies. Best results are shown in bold.}
\label{tab:ablation_pretrain_alt}
\footnotesize
\renewcommand{\arraystretch}{1.15}
\setlength{\tabcolsep}{0pt}
\setlength{\aboverulesep}{0.1ex}
\setlength{\belowrulesep}{0pt}
\setlength{\cmidrulesep}{0.1ex}

\begin{tabular}{@{}
    >{\centering\arraybackslash}p{0.22\columnwidth}
    >{\centering\arraybackslash}p{0.15\columnwidth}
    >{\centering\arraybackslash}p{0.20\columnwidth}
    >{\centering\arraybackslash}p{0.21\columnwidth}
    >{\centering\arraybackslash}p{0.22\columnwidth}
@{}}
\toprule
\rowcolor{mmhead}
& &
\textbf{Classification} &
\textbf{Detection} &
\textbf{Segmentation} \\
\cmidrule{3-5}

\rowcolor{mmsubhead}
\textbf{Masking} &
\textbf{Data} &
\textbf{SFP} &
\textbf{FIS} &
\textbf{PUBSEG} \\
\cmidrule{3-5}

\rowcolor{mmmetric}
& &
\textit{Acc} &
\textit{mAP@50:95} &
\textit{mDice} \\
\midrule
MAE          & 430K & 96.06          & 53.0          & 85.11 \\
HOG+Gaussian & 260K & 95.98          & 53.4         & 90.49  \\
\rowcolor{mmrowhi}
HOG+Polar    & 260K & \textbf{96.46} & \textbf{54.4} & \textbf{99.59} \\
\bottomrule
\end{tabular}
\end{table}
PVSS is designed to mitigate the substantial visual and semantic redundancy caused by continuous scanning. Unlike a random reduction of the pre-training set, the full method uses an informative subset obtained through dual-stage de-duplication. The results show that, after introducing PVSS, the model still achieves better downstream performance on a more compact pre-training set. This finding indicates that, for ultrasound image pre-training, increasing the effective density of training samples is more important than simply enlarging the data scale. 

ABRC targets the large invalid background regions that are common in ultrasound images. By explicitly constraining the effective imaging region, ABRC focuses the pre-training process on anatomically meaningful areas. The results show that introducing ABRC leads to more consistent gains on structure-sensitive tasks, which indicates that suppressing background interference helps learn representations with stronger spatial specificity.


PTCM addresses the non-uniform distribution of information in ultrasound images, where different patches do not contribute equally to representation learning. By combining a polar prior with local gradient information, PTCM guides the model to prioritize regions with higher information content. The results show that, after introducing PTCM, the model further improves on detection and segmentation tasks, which suggests that this strategy more effectively models local structures, boundary regions, and diagnostically relevant targets.

Overall, Table~\ref{tab:ablation_pretrain} shows that PVSS, ABRC, and PTCM address redundancy, invalid background regions, and non-uniform information distribution in ultrasound images, respectively, and together constitute the main source of the performance gains of PolarMAE. This result indicates that the advantage of the proposed method does not arise from a local improvement of any single module, but from a coordinated design tailored to the characteristics of ultrasound imaging.

\begin{table}[!t]
\centering
\caption{Ablation study on different pre-training strategies. Best results are shown in bold. SFP denotes Six Fetal Planes, FIS denotes Fetus Intracranial Structure, and FHB denotes Fetal Head Biometry.}
\vspace{-5pt}
\label{tab:ablation_pretrain}
\footnotesize
\renewcommand{\arraystretch}{1.12}
\setlength{\tabcolsep}{0pt}
\setlength{\aboverulesep}{0pt}
\setlength{\belowrulesep}{0pt}
\setlength{\cmidrulesep}{0.1ex}

\begin{tabular}{@{}
    >{\centering\arraybackslash}p{0.25\columnwidth}
    >{\centering\arraybackslash}p{0.15\columnwidth}
    >{\centering\arraybackslash}p{0.18\columnwidth}
    >{\centering\arraybackslash}p{0.23\columnwidth}
    >{\centering\arraybackslash}p{0.22\columnwidth}
@{}}
\toprule
\rowcolor{mmhead}
& &
\textbf{Classification} &
\textbf{Detection} &
\textbf{Segmentation} \\
\cmidrule{3-5}

\rowcolor{mmsubhead}
\textbf{Model} &
\textbf{Data} &
\textbf{SFP} &
\textbf{FIS} &
\textbf{PUBSEG} \\
\cmidrule{3-5}

\rowcolor{mmmetric}
& &
\textit{Acc} &
\textit{mAP@50:95} &
\textit{mDice} \\
\midrule
MAE      & 430K & 96.06          & 53.0          & 85.11 \\
PVSS  & 260K & 96.06          & 53.0          & 92.39 \\
PVSS+ABRC     & 260K & 96.14          & 53.3          & 93.88 \\
\rowcolor{mmrowhi}
PVSS+ABRC+PTCM    & 260K & \textbf{96.46} & \textbf{54.4} & \textbf{99.59} \\
\bottomrule
\end{tabular}
\end{table}

\subsubsection{Ablation on Masking Strategy}
\label{subsubsec:mask_ablation}

To evaluate the effect of masking strategy on self-supervised pre-training for ultrasound images, we further compare three patch selection schemes, as reported in Table~\ref{tab:ablation_pretrain_alt}: the random masking strategy used in standard MAE, a strategy that combines HOG with a Gaussian distribution, and the HOG and polar-coordinate joint strategy used in our method. This experiment examines two questions. The first is whether masking design should explicitly model the non-uniform distribution of informative content in ultrasound images. The second is whether, after introducing structure-aware guidance, polar-coordinate constraints further improve adaptation to the imaging characteristics of ultrasound.

The results show that HOG+Polar achieves the best performance on all downstream tasks, which indicates that masking strategy directly affects the modeling of diagnostically relevant regions during pre-training. Compared with the random masking strategy of standard MAE, the proposed method guides the model more effectively toward regions with richer structural information and higher diagnostic value, thereby improving representation quality. This finding suggests that self-supervised pre-training for ultrasound images should not directly follow the random masking paradigm developed for general vision tasks, but should instead explicitly account for the non-uniform spatial distribution of informative content.

A further comparison between HOG+Gaussian and HOG+Polar shows that the latter consistently performs better. This result indicates that local structural information alone is insufficient to fully characterize the distribution of informative regions in ultrasound images. With polar-coordinate constraints, the masking strategy better matches the spatial organization of ultrasound imaging. Therefore, the advantage of the proposed method arises not only from the use of local structural cues, but also from integrating patch selection with the geometric priors of ultrasound imaging.
\begin{figure}[!t]
  \centering
  \includegraphics[width=\columnwidth]{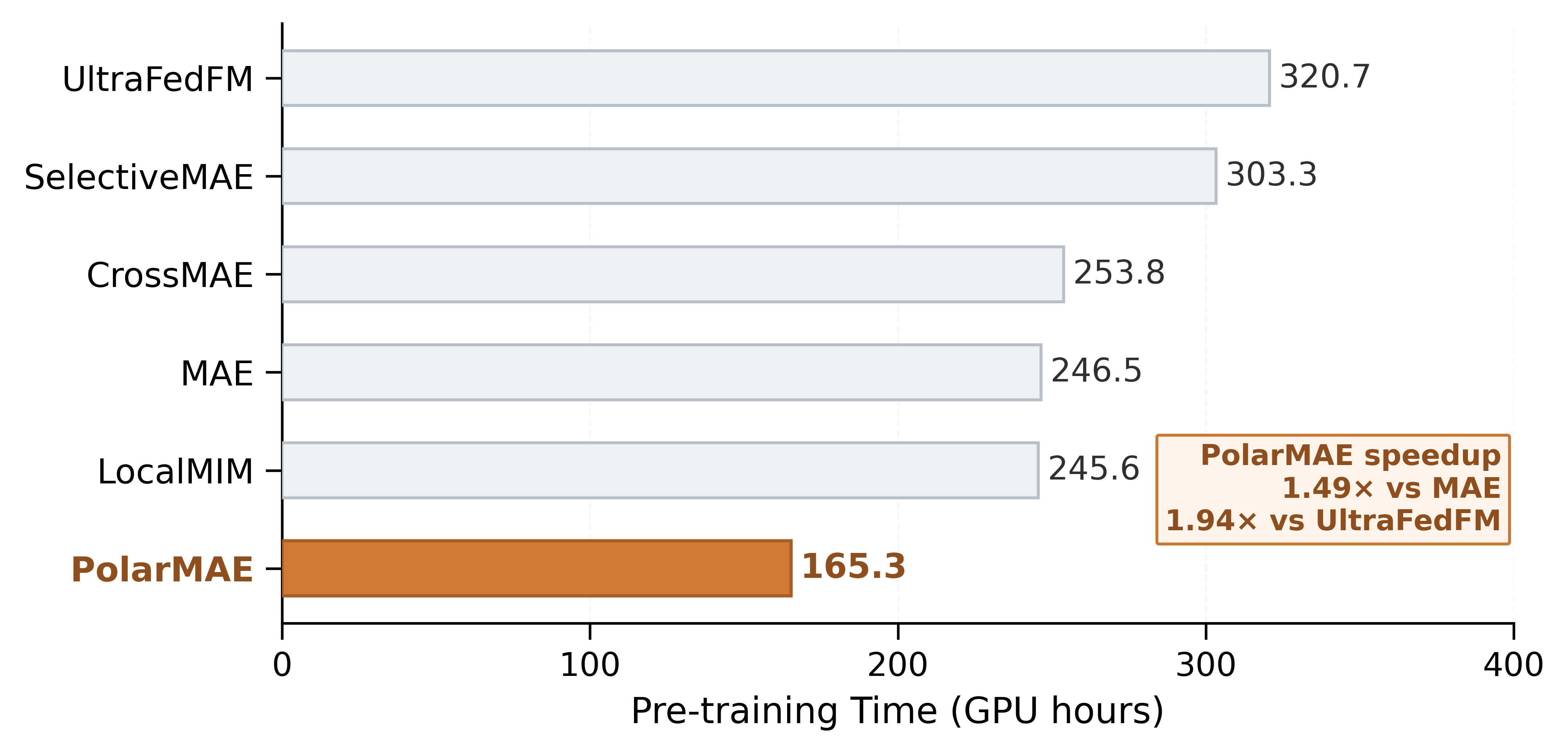}
  \caption{Pre-training efficiency comparison across methods. PolarMAE achieves the best training efficiency with the lowest pre-training cost.}
  \Description{Pre-training efficiency comparison for ultrasound self-supervised learning, showing that PolarMAE achieves the best training efficiency with the lowest pre-training cost.}
  \vspace{-13pt}
  \label{fig:pretrain_efficiency}
\end{figure}

\subsection{Computational Efficiency (RQ3)}
To answer RQ3 and analyze whether our proposed method can maintain superior representation quality while achieving highly efficient training, we comprehensively compare the pre-training time of PolarMAE against various state-of-the-art (SOTA) methods.

As illustrated in Fig.~\ref{fig:pretrain_efficiency}, we observe that PolarMAE achieves the most optimal efficiency among all evaluated models without compromising its state-of-the-art performance. Specifically, our method requires only 165.3 GPU hours for pre-training. Compared to the most competitive baseline regarding efficiency, LocalMIM (245.6 GPU hours), PolarMAE achieves a notable speedup of approximately 1.48$\times$, striking a superior balance between computational economy and representation quality.

\begin{figure}[!t]
  \centering
  \includegraphics[width=\columnwidth]{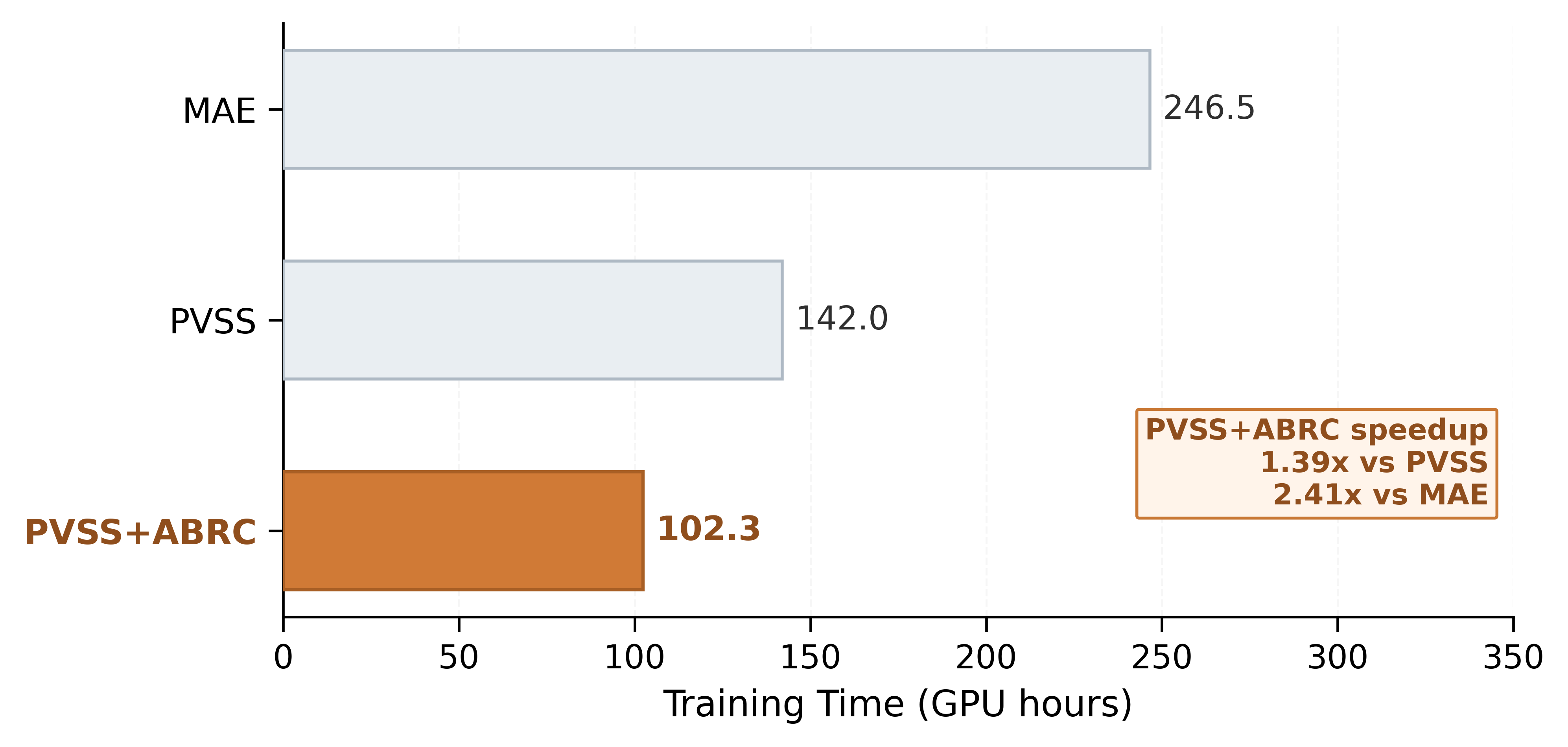}
  \caption{Training efficiency with progressive module introduction. Using MAE as the baseline, adding PVSS and then PVSS+ABRC steadily reduces the pre-training cost.}
  \label{fig:ablation_efficiency}
\end{figure}

To further investigate the underlying drivers of this efficiency, we analyze the individual contributions of the Progressive Visual-Semantic Screening (PVSS) module and the Acoustic-Bounded Region Constraint (ABRC) module on pre-training acceleration. As depicted in Fig.~\ref{fig:ablation_efficiency}, integrating the PVSS module for semantic screening significantly reduces the pre-training time from the baseline MAE’s 246.5 GPU hours down to 142.0 GPU hours. Subsequently, incorporating the ABRC module yields an additional acceleration, further decreasing the required computational time from 142.0 to a mere 102.3 GPU hours. This substantial reduction—culminating in an overall 2.41$\times$ speedup compared to the baseline—solidly demonstrates the necessity and effectiveness of the proposed PVSS and ABRC modules in accelerating the unsupervised pre-training process.

\subsection{Data Scalability (RQ4)}
To evaluate the data scalability of our pre-training framework, we conducted a series of experiments across varying data scales. Due to space constraints, the detailed analysis and results are provided in the Supplementary Material.

\section{Conclusion}
\label{sec:conclusion}
In this paper, we introduce and investigate an efficient unsupervised pre-training framework, PolarMAE, explicitly tailored for fetal ultrasound (US) analysis. We empirically show that existing Masked Image Modeling (MIM) methods are inefficient due to their inability to handle US-specific continuous scanning redundancy and unique acoustic physical characteristics. To address this, PolarMAE incorporates a Progressive Visual-Semantic Screening (PVSS) and an Acoustic-Bounded Region Constraint (ABRC) to eliminate invalid computations, alongside a Polar-Texture Collaborative Masking (PTCM) that leverages both the physical prior of acoustic beamforming and local texture. From our extensive experiments across diverse downstream tasks, we mainly conclude that: 1) filtering redundant ultrasound images and restricting reconstruction strictly to valid fan-shaped anatomical regions significantly accelerate the pre-training process while preventing overfitting to useless backgrounds; 2) it is critical to inject the physical prior of acoustic beamforming and collaborative texture information to capture the underlying non-uniform radial imaging patterns and prioritize essential tissue structures. In future work, we plan to extend this physics-aware framework to multimodal ultrasound analysis—such as integrating US images with clinical text reports—to further unleash the potential of foundation models in prenatal diagnosis.

\bibliographystyle{ACM-Reference-Format}
\bibliography{ref}

\end{document}